\pdfoutput=1

\documentclass[11pt]{article}

\usepackage[final]{acl}

\usepackage{colortbl}
\usepackage{amsmath}
\usepackage{booktabs}
\usepackage{multirow}
\usepackage{times}
\usepackage{latexsym}
\usepackage[T1]{fontenc}

\usepackage[utf8]{inputenc}

\usepackage{microtype}

\usepackage{inconsolata}

\usepackage{graphicx}

\usepackage{amsfonts}

\usepackage{algorithm}
\usepackage{algpseudocode}
\usepackage{tabularx}

\usepackage{threeparttable}

\usepackage{pifont}
\newcommand{\cmark}{\textcolor{green!70!black}{\ding{51}}} 
\newcommand{\xmark}{\textcolor{red}{\ding{55}}}            

\title{Beyond Demonstrations: \\ Dynamic Vector Construction from Latent Representations}

\author{
    Wang Cai$^{1,2,3}$\thanks{~~Equal contribution.}, 
    Hsiu-Yuan Huang$^{1,2,4}$\footnotemark[1],
    Zhixiang Wang$^{1,2,3}$, 
    Yunfang Wu$^{1,2,4}$\thanks{~~Corresponding author.} \\
    $^{1}$National Key Laboratory for Multimedia Information Processing, Peking University \\ 
    $^{2}$MOE Key Laboratory of Computational Linguistics, Peking University \\
    $^{3}$School of Software and Microelectronics, Peking University \\
    $^{4}$School of Computer Science, Peking University \\
    \texttt{\{caiwang, huang.hsiuyuan, ekko\}@stu.pku.edu.cn}, 
    \texttt{wuyf@pku.edu.cn}
}

\begin{document}
\maketitle
\begin{abstract}
In-Context derived Vector (ICV) methods extract task-relevant representations from large language models (LLMs) and reinject them during inference, achieving comparable performance to few-shot In-Context Learning (ICL) without repeated demonstration processing. 
However, existing ICV methods remain sensitive to ICL-specific factors, often use coarse or semantically fragmented representations as the source of the vector, and rely on heuristic-based injection positions, limiting their applicability.
To address these issues, we propose \textbf{Dynamic Vector} (DyVec), which incorporates an \textit{Exhaustive Query Rotation} (EQR) strategy to extract robust semantically aggregated latent representations by mitigating variance introduced by ICL. It then applies \textit{Dynamic Latent Segmentation and Injection} to adaptively partition representations based on task complexity and leverages REINFORCE-based optimization to learn optimal injection positions for each segment.
Experiments results show that DyVec outperforms few-shot ICL, LoRA, and prior ICV baselines. Further analysis highlights the effectiveness of dynamically segmenting and injecting semantically aggregated latent representations. DyVec provides a lightweight and data-efficient solution for inference-time task adaptation.

\end{abstract}

\section{Introduction}

\begin{figure}[!t]
    \centering
    \setlength{\tabcolsep}{0.8mm}
      \includegraphics[width=\linewidth]{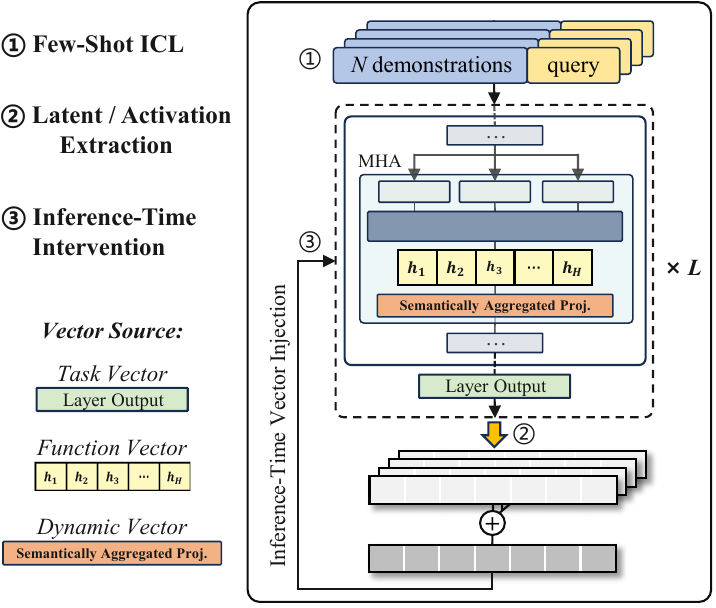}
      \caption {General pipeline of In-Context derived Vector (ICV) methods, illustrating how task-specific representations are extracted from LLMs during few-shot ICL to construct vectors, which are then injected back into frozen LLMs for inference-time intervention and task adaptation. These representations can be either raw activations (e.g., attention heads) or more abstract latent states (e.g., transformer layer outputs).}
    \label{fig:teaser}
\end{figure}

Large Language Models (LLMs) have demonstrated emergent capabilities in few-shot learning, allowing them to perform new tasks by conditioning on just a few demonstrations in the input prompt—without any parameter updates \cite{brown2020languagemodelsfewshotlearners}. This paradigm is known as In-Context Learning (ICL).
Despite its success in low-resource settings, it incurs substantial computational overhead, as each inference requires repeatedly encoding lengthy prompts with demonstration examples. This inefficiency hinders its scalability in real-world applications.

Recent studies have introduced In-Context derived Vector (ICV) methods~\cite{hendel2023_ICL_TV, todd2024functionvectorslargelanguage}, which extract internal activations or latent representations of LLM that capture task-specific information during ICL inference. These vectors can then be injected into LLMs at inference time to approximate few-shot performance~\cite{hojel2024findingvisualtaskvectors}, as illustrated in Figure~\ref{fig:teaser}.
ICV methods have shown promise in various applications—such as promoting honesty~\cite{NEURIPS2023_81b83900}, reducing harmful outputs~\cite{liu2024incontextvectorsmakingcontext}, and enabling role-playing~\cite{potertì2025designingrolevectorsimprove}—all while maintaining the efficiency of zero-shot inference.

While ICV presents a promising alternative to standard few-shot ICL, it still suffers from three key limitations: 
(1) Existing methods construct vectors from sources that are either too coarse (e.g., transformer layer outputs) or too semantically isolated (e.g., the raw activations of independent attention heads). 
Both strategies fail to fully capture rich task semantics, resulting in performance that lags behind few-shot ICL.
(2) Prior methods improve robustness by averaging the extracted latent representations from repeated inferences, partially mitigating ICL sensitivity. However, this strategy offers limited gains and fails to fundamentally resolve the sensitivity inherent to ICL.
(3) Finally, selecting the optimal vector injection location often requires extensive validation data and exhaustive searches across layers or heads, making the process both inefficient and impractical.

To tackle these challenges, we propose Dynamic Vector (DyVec)—a inference-time intervention method designed to directly address these three limitations:

(1) To overcome the coarse or semantically isolated nature of prior representations, DyVec uses the semantically aggregated projections in Multi-Head Attention (MHA) as the source for vector construction, enabling the model to capture richer task semantics through inter-head interactions.

(2) To enhance robustness against ICL sensitivity, we introduce \textit{Exhaustive Query Rotation} (EQR), which systematically rotates the query position within a fixed set of demonstrations and aggregates the extracted latent representations, providing a more stable representation to construct vector.

(3) To eliminate reliance on validation data and heuristics, we propose \textit{Dynamic Latent Segmentation and Injection}, which adaptively partitions the extracted representations based on task complexity and resource constraints. We further employ REINFORCE to learn optimal injection positions for each segment in a data-driven manner.

Finally, the constructed Dynamic Vectors are utilized to perform \textit{Inference-Time Intervention} by injecting them back into the LLM for task adaptation. Empirical results demonstrate that our three strategies not only outperform standard few-shot ICL under comparable inference costs, but also achieve superior results compared to existing ICV methods. Table~\ref{tab:compare_methods} provides a comparison between DyVec and existing ICV-based methods. 

\begin{table*}[t]
\centering
\small
\begin{threeparttable}
\begin{tabular}{lcccccc}
\toprule
~
& \multicolumn{2}{c}{\textbf{Dynamism}} 
& \multicolumn{4}{c}{\textbf{Vector Construction}} \\
\cmidrule(lr){2-3} \cmidrule(lr){4-7}
\textbf{Method}
& \textbf{Seg.} 
& \textbf{Inj.} 
& \textbf{Source} 
& \textbf{Computation} 
& \textbf{Granularity} 
& \textbf{Injection Position} \\
\midrule
TV & \xmark & \xmark & Transformer Layer Output & Avg. & Layer $L$ & Valid. \\
FV & \xmark & \xmark & Attention Head Activation & Avg. & Attention Head $H$ & CIE + Valid. \\
DyVec & \cmark & \cmark & Semantically Aggregated Latent & EQR + Avg. & Dynamic Segment $S$ & REINFORCE \\
\bottomrule
\end{tabular}
\vspace{0.2em}
{\scriptsize\raggedright
Seg. = Segmentation, Inj. = Injection, Avg. = Averaging, 
EQR = Exhaustive Query Rotation, CIE = Causal Indirect Effect, 
Valid. = Validation-Driven Layer Search.\par}
\caption{Comparison of DyVec with existing ICV methods across key aspects of vector construction.}
\label{tab:compare_methods}
\end{threeparttable}
\end{table*}

Our contributions are summarized as follows:
\begin{itemize}
    \item We propose Dynamic Vector, a novel ICV-based inference-time intervention method that injects task-specific vectors into frozen LLMs, achieving few-shot performance while retaining the efficiency of zero-shot inference.
    \item  We introduce Exhaustive Query Rotation and a dynamic segmentation strategy with REINFORCE-based injection, enabling the extraction of robust task representations and their flexible integration into the model’s latent space to support vector construction.
    \item We conduct extensive experiments across multiple tasks, demonstrating the effectiveness and generality of our approach.
\end{itemize}




    

\section{Related Work}
\paragraph{In-Context Learning (ICL)}
ICL enables LLMs to perform new tasks without parameter updates by conditioning on a few task-specific examples in the input prompt~\cite{brown2020languagemodelsfewshotlearners}, and has been extended to various applications~\cite{wei2023chainofthoughtpromptingelicitsreasoning,wang2023selfconsistencyimproveschainthought,yao2023reactsynergizingreasoningacting}.
However, ICL faces two major challenges:
(1) \textit{Inefficiency} — each inference involves a full forward pass over lengthy prompts with repeated demonstrations, leading to high memory and compute costs, especially in resource-limited settings \cite{liu2022fewshotparameterefficientfinetuningbetter};
(2) \textit{Instability} — performance is highly sensitive to prompt design, including example order and selection~\cite{liu2021makesgoodincontextexamples,rubin2022learningretrievepromptsincontext}.



\paragraph{In-Context Vector (ICV)}
Transformers encode the semantics of ICL demonstrations within their internal activations~\cite{hendel2023_ICL_TV, todd2024functionvectorslargelanguage, rimsky2024_steering,  huang2024multimodal}. These activations, termed in-context vectors, capture task-specific signals and can be injected at inference time to emulate few-shot behavior in a zero-shot setting—achieving comparable performance while retaining the efficiency of zero-shot inference.

ICVs have been applied to guide model behavior across tasks, such as promoting honesty~\cite{NEURIPS2023_81b83900}, reducing harmful outputs~\cite{panickssery2024steeringllama2contrastive, liu2024incontextvectorsmakingcontext, Wang_2025_CVPR}, or enabling role-playing~\cite{potertì2025designingrolevectorsimprove}. ICVs have also been explored in vision~\cite{hojel2024findingvisualtaskvectors} and multimodal models~\cite{huang2024multimodal}, demonstrating broad applicability.

However, prior ICV-based methods often fail to outperform standard ICL~\cite{todd2024functionvectorslargelanguage}, or require extra training data~\cite{huang2024multimodal}. In contrast, our proposed DyVec improves over few-shot ICL without any additional data.

\section{Preliminary}
In ICL, LLMs are prompted with a few input-output demonstrations followed by a query. Formally, an ICL prompt is defined as:
\begin{equation}
\mathcal{P} = {(x_1, y_1), \ldots, (x_K, y_K), x^{\text{query}}}
\end{equation}
where the model processes $\mathcal{P}$ in a single forward pass and generates $y^{\text{query}}$ as the prediction.

Recent work has shown that LLMs encode task-specific signals from ICL demonstrations into their hidden states \cite{hendel2023_ICL_TV, todd2024functionvectorslargelanguage, huang2024multimodal}. These representations, which we refer to as ICVs, can be extracted and reused to induce few-shot-like behavior during zero-shot inference. 

Most existing ICV methods \cite{todd2024functionvectorslargelanguage, huang2024multimodal} construct vectors from raw attention head activations. Specifically, for attention head $j$ in layer $i$, the output is given by:
\begin{equation}
\mathbf{A}_{(i,j)} = \text{Softmax}\left(\frac{\mathbf{Q}_{(i,j)}\mathbf{K}^{\top}_{(i,j)}}{\sqrt{d_h}}\right)\mathbf{V}_{(i,j)}
\label{eq:attention}
\end{equation}
During inference, the extracted ICV $\mathbf{A}_{(i,j)}$ is injected back into the same location in the frozen model, scaled by a factor $\beta$:
\begin{equation}
\mathbf{\hat{A}}_{(i,j)} = \mathbf{A}_{(i,j)}^{\text{infer}} + \beta \cdot \mathbf{A}_{(i,j)},
\end{equation}
This injection shifts the model’s internal activations in a task-specific direction, effectively adapting the model without demonstrations in the prompt.

The performance of ICVs is primarily determined by two critical aspects: 
(1) the quality of the extracted representation to construct vectors, and (2) the informativeness of the positions in the model where these vectors are injected.

Existing approaches typically extract ICVs by running ICL with randomly sampled demonstrations and averaging the resulting representations. While this mitigates the model’s sensitivity to factors such as demonstration position and composition, it does not fully resolve the issue. Moreover, prior methods construct vectors directly from raw attention head outputs $A$, overlooking potential inter-head interactions that capture richer semantic information. To determine the optimal injection locations for these vectors, existing approaches either rely on validation performance \cite{hendel2023_ICL_TV}, or incur significant computational overhead by exhaustively searching across tasks \cite{todd2024functionvectorslargelanguage}.
To address these issues, we propose DyVec.

\section{Dynamic Vector Construction from Latent Representations}

\begin{figure*}[!t]
    \centering
    \setlength{\tabcolsep}{0.8mm}
      \includegraphics[width=0.95\textwidth]{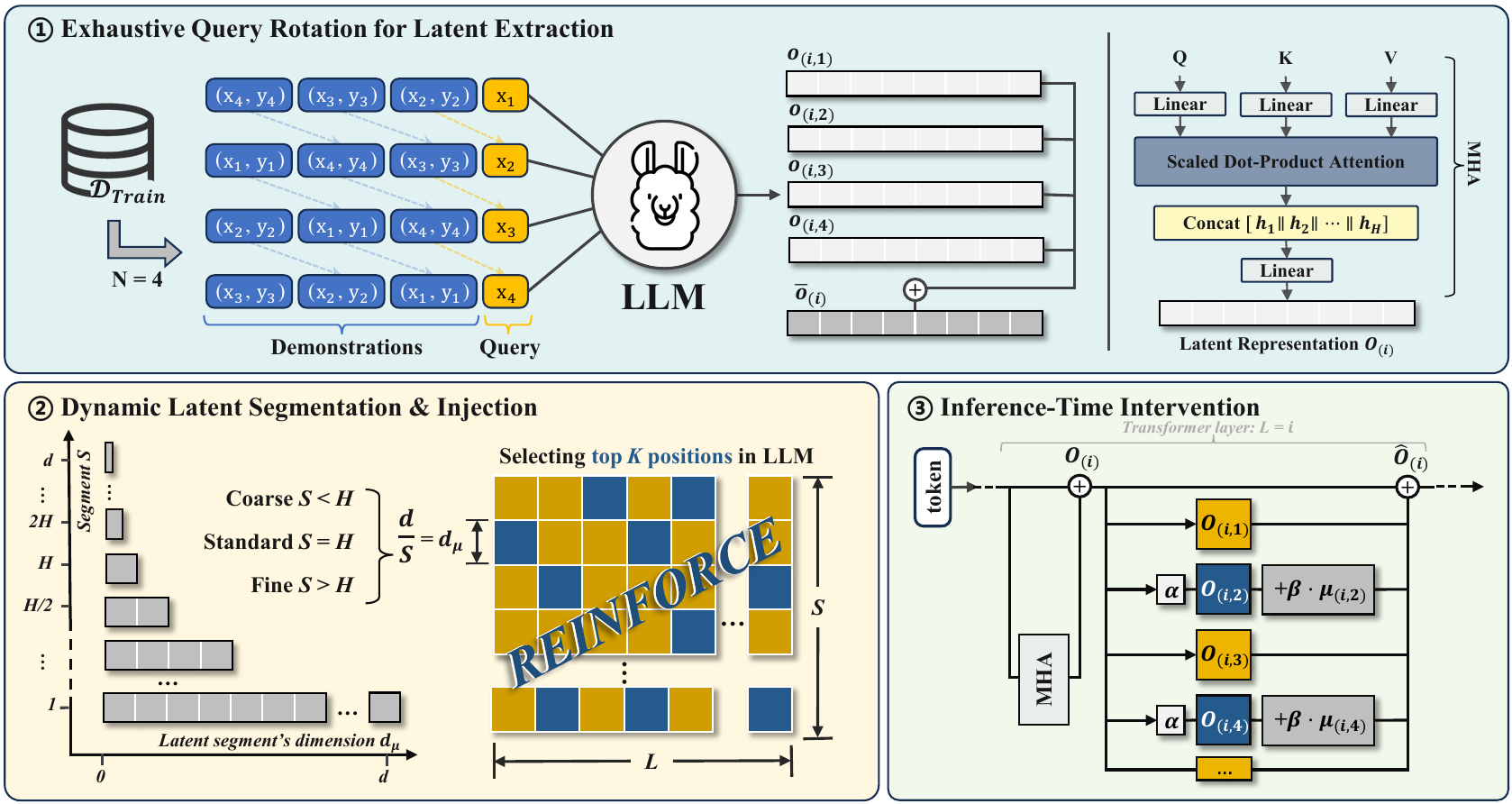}
      \caption {The overview of our proposed model.}
    \label{fig:pipeline}
\end{figure*}

Figure~\ref{fig:pipeline} provides an overview of the complete DyVec pipeline, which consists of three key stages: (1) an \textit{Exhaustive Query Rotation} strategy for extracting robust, semantically aggregated latent representations, (2) \textit{Dynamic Segmentation and Injection} for constructing vectors, and (3) \textit{Inference-Time Intervention} via vector injection.

\subsection{Exhaustive Query Rotation for Robust Latent Extraction}
\label{sec:task_vector_extraction}

To enhance robustness against biases introduced by the order and composition of demonstrations, we propose an \textit{Exhaustive Query Rotation} (EQR) strategy for reliable and semantically aggregated latent representation extraction. Given a sampled subset of $N$ labeled instances from the training set, we systematically rotate through each instance by treating it once as the query and using the remaining $N-1$ examples as demonstrations. This results in $N$ distinct ICL prompts:
\begin{equation}
\small
\mathcal{P}_n = \left\{(x_m, y_m) \mid m \neq n,\ n \in \{1, 2, ..., N\} \right\} \cup \{x_n\}
\end{equation}

For each prompt $\mathcal{P}_n$, we perform a forward pass through the model. At the last token position of the prompt, we extract latent representations from the semantically aggregated projections within the MHA modules across all transformer layers. The latent representation of layer $i$ is denoted as:
\begin{equation}
\begin{aligned}
\mathbf{A}_{(i)} &= [\mathbf{A}_{(i,1)} \| \cdots \| \mathbf{A}_{(i,H)}] \\
\mathbf{O}_{(i)} &= \mathbf{A}_{(i)} \mathbf{W}^O_{(i)}
\label{eq:proj_output}
\end{aligned}
\end{equation}
where $\mathbf{W}^O_{(i)} \in \mathbb{R}^{d \times d}$ is the output projection matrix at layer $i$, and $\mathbf{o}_{(i,n)}$ denotes the latent representation at the last token position for layer $i$ and prompt $n$.
We then average the latent representations extracted from all $N$ rotated prompts to obtain a robust representation, which will serves as the basis for vector construction:
\begin{equation}
\overline{\mathbf{o}}_{(i)} = \frac{1}{N}\sum_{n=1}^{N} \mathbf{o}_{(i,n)}
\end{equation}
The EQR strategy produces a stable and sementic aggregated task-specific latent representation by ensuring each instance contributes equally. However, using the entire $\overline{\mathbf{o}}_{(i)}$ as a monolithic vector is overly rigid. To introduce greater task-specific flexibility, we propose a dynamic segmentation and injection mechanism that enables more fine-grained, task-centric control.

\subsection{Dynamic Latent Segmentation and Injection}
\label{sec:optimization}

To extract task-specific information in a more dynamic manner, we re-partition the latent representation $\overline{\mathbf{o}}_{(i)}$ into $S$ segments, where $S \in \{s \mid d \bmod s = 0\}$ denotes the set of values that evenly divide the hidden dimension $d$ of the LLM. We obtain a list of latent segments, denoted as:
\begin{equation}
\mu_{(i)} = \left[ \mu_{(i,1)}, \mu_{(i,2)}, \cdots, \mu_{(i,S)} \right]
\end{equation}
Each dynamic latent segment $\mu_{(i,j)} \in \mathbb{R}^{d_\mu}$ represents a contiguous slice of the latent representation $\overline{\mathbf{o}}_{(i)}$, where the dimensionality of each segment is $d_\mu = d / S$.

When $S = H$ (i.e., the number of attention heads in the model), this segmentation corresponds to a \textit{standard} split. Increasing $S$ leads to \textit{finer-grained} segments with smaller $d_\mu$, enabling more detailed inspection of attention behavior across more localized subspaces; decreasing $S$ yields \textit{coarser} segments with larger $d_\mu$, potentially reducing computational cost and still preserve the key task semantics. Notably, when $S=1$, the $\mu_{(i)}$ degenerates into a layer-shape representation. This segmentation flexibility allows DyVec to adjust the granularity of vector construction according to the task’s complexity and the available computational resources.

By segmenting $\overline{\mathbf{o}}_{(i)}$, we capture how task signals are distributed across layers and subspaces. However, not all positions—indexed by layer $i$ and segment $j$—contribute equally to task-specific behavior. To identify the most informative positions in the latent space, we introduce the notion of the \textit{Dynamic Vector}: a selected set of latent representations that collectively represent the task-specific distribution. Formally, we aim to select an optimal subset \( \mathcal{Y}^* \subseteq \{(i,j)\} \), and construct the \textit{Dynamic Vector} as:
\begin{equation}
\theta = \{\mu_{(i,j)} \mid (i,j) \in \mathcal{Y}^*\}
\label{eq:task_vector}
\end{equation}

We define an intervention function \( \mathcal{L} \), which injects the \textit{Dynamic Vector} \( \theta \) into the corresponding positions during zero-shot inference, thereby modifying the model’s output distribution. The function \( \mathcal{L} \) operates as a linear intervention strategy, combining the selected latent segment $\mu_{(i,j)}$ with the model’s native activations via weighted summation.

To discover the most effective intervention positions, we adopt REINFORCE~\cite{williams1992reinforce}, a policy gradient method from reinforcement learning.
The optimization over the candidate set of positions $L \times S$ is detailed in Algorithm~\ref{alg:REINFORCE}. Specifically, 
we first parameterize a Bernoulli distribution over all possible insertion positions, where each element corresponds to a latent segment at a given layer. At each optimization step, binary masks are sampled from this distribution to determine which segments are activated for constructing the dynamic vector \( \theta \). This vector is then injected into the model using strategy \( \mathcal{L} \), and the model output \( \mathcal{L}(\mathcal{M}, \theta) \) is evaluated via cross-entropy loss on the training set \( \mathcal{D}_{\text{train}} \). The reward \( R \) is defined as the negative of this loss. We update the Bernoulli parameters using the REINFORCE algorithm to encourage selection of positions that improve downstream performance. We also apply \( \text{Clip}(x, \epsilon) \) to keep each parameter within the interval \( [\epsilon, 1 - \epsilon] \).

\begin{algorithm}[t]
\small
\caption{REINFORCE}
\label{alg:REINFORCE}
\begin{algorithmic}[1]
\Require Positions $L \times S$, learning rate $\alpha_p$, steps $T$
\State Initialize $p_{i,j} \gets 0.5$, $\forall(i,j)$
\For{$t = 1$ to $T$}
    \State Sample $m_{i,j} \sim \text{Bernoulli}(p_{i,j})$
    \State $\theta \gets \left\{ \mu_{i,j} \mid m_{i,j} = 1 \right\}$

    \State $R \gets -\text{CE}(\mathcal{L}(\mathcal{M}, \theta), \mathcal{D}_{\text{train}})$
    \For{$(i,j)$}
        \State \makebox[\dimexpr\linewidth-2em][l]{%
            $
            p_{i,j} \gets \text{Clip}\left(
            p_{i,j} + \alpha_p \cdot \frac{m_{i,j} - p_{i,j}}{p_{i,j}(1-p_{i,j})} \cdot R,\epsilon
            \right)
            $
        }
    \EndFor
\EndFor
\State $\mathcal{Y}^* \gets \text{TopK}(p_{i,j},\ \lceil \sum p_{i,j} \rceil)$
\State \Return $\{\mu_{i,j} : (i,j) \in \mathcal{Y}^*\}$
\end{algorithmic}
\end{algorithm}


Notably, this optimization process does not require any additional training data. All queries and corresponding labels in $\mathcal{D}_{\text{train}}$ are drawn directly from the original example set, allowing DyVec to operate effectively even under extremely limited supervision.
In the next section, we elaborate on how to select the optimal intervention function \( \mathcal{L} \) from a set of candidate strategies.

\subsection{Dynamic Vector Injection for Inference-Time Intervention}
\label{sec:intervention_strategy}
During zero-shot inference, for each position \( (i,j) \in \mathcal{Y}^* \), we modify the latent representaion \( O_{(i,j)} \) by injecting the corresponding DyVec segment scaled by a strength factor $\beta$:
\begin{equation}
\hat{O}_{(i,j)} = \alpha \cdot O_{(i,j)} + \beta \cdot \mu_{(i,j)}
\end{equation}
Here, $\alpha \in \{0,1\}$ acts as a binary gating mechanism that dynamically controls whether the original output is retained, while $\beta$ modulates the influence of the injected dynamic latent segment. We explore a small set of such intervention strategies and select the one that yields the lowest cross-entropy loss on the training data.

Through the above three steps, our method is able to automatically construct and inject DyVec into the model using a limited number of ICL demonstrations, without relying on additional training data or updating any model parameters. This significantly enhances the generalization ability of large language models in the zero-shot setting.

\section{Experimental Setup}
We evaluate DyVec on three 7B-scale open-source LLMs: LLaMA-2-7B-Chat~\cite{touvron2023llama2openfoundation}, Qwen2-7B~\cite{yang2024qwen2technicalreport}, and DeepSeek-LLM-7B-Chat~\cite{deepseekai2024deepseekllmscalingopensource}.

Our main experiments focus on \textbf{six classification tasks} covering sentiment analysis, sarcasm detection, medical relation extraction, and topic/question classification, using datasets such as NHSD~\cite{misra2022newsheadlinesdatasetsarcasm}, Sarcasm~\cite{nikesh66_sarcasm_dataset}, SST2~\cite{socher-etal-2013-recursive}, ADE~\cite{GURULINGAPPA2012885}, AG\_News~\cite{zhang2016characterlevelconvolutionalnetworkstext}, and TREC6~\cite{trec6_hf}.We randomly sample 1,000 instances per task to form the test set for evaluation.

In addition, we include \textbf{six generation tasks} adapted from~\citet{todd2024functionvectorslargelanguage}, involving lexical and grammatical transformations (e.g., Antonym, Capitalize, English-French). These tasks are used to assess generalization beyond classification but are not the primary focus of this work. Full results are reported in Table~\ref{tab:linguistic task}.

To evaluate the effectiveness of our proposed method DyVec, we conduct comparisons in two directions: (1) against standard adaptation baselines, including both non-trainable approaches such as \textbf{Few-Shot ICL}, and trainable paradigms such as \textbf{LoRA} and \textbf{Linear Probing (LP)}; and (2) against ICV methods, including \textbf{Task Vector (TV)} and \textbf{Function Vector (FV)}.

Further implementation details are provided in Appendix~\ref{sec:prompt construction} (Prompt Construction Details), Appendix~\ref{sec:Data Construction Details} (Data Construction Details), Appendix~\ref{sec:baselines} (Baselines), and Appendix~\ref{sec:Method} (DyVec intervention strategy).

\section{Results and Analysis}

\begin{table*}[!t]
\centering
\renewcommand{\arraystretch}{1.0}
\footnotesize
\setlength{\tabcolsep}{6pt}

\begin{tabular}{c c c c c c c c c |>{\columncolor[gray]{0.9}}c}
\toprule
\textbf{Model} & \textbf{Data size} & \textbf{Method} & NHSD & Sarcasm & SST2 & ADE & AG\_News & TREC6 & \textbf{Average} \\
\midrule

\multirow{12}{*}{LLaMA} 
& \multirow{4}{*}{4} & ICL  & 34.80 & 50.70 & 90.20 & 52.60 & 45.80 & 21.90 & 49.33 \\
&                     & LoRA & 47.10 & 27.90 & 43.40 & 47.00 & 40.30 & 16.70 & 37.07 \\
&                     & LP   & 60.30 & 59.10 & 50.20 & 52.40 & 44.20 & 22.70 & 48.15 \\
&                     & DyVec& 48.50 & 57.40 & 55.60 & 56.90 & 64.20 & 26.40 & \textbf{51.50} \\
\cmidrule(lr){2-10}
& \multirow{4}{*}{8} & ICL  & 24.00 & 64.90 & 93.10 & 57.10 & 71.20 & 24.10 & 55.73 \\
&                    & LoRA & 41.90 & 59.70 & 81.00 & 55.00 & 46.80 & 31.30 & 52.62 \\
&                    & LP   & 66.90 & 61.60 & 72.50 & 59.20 & 52.70 & 33.40 & 57.72 \\
&                    & DyVec& 49.00 & 50.40 & 91.50 & 67.40 & 79.20 & 25.40 & \textbf{60.48} \\
\cmidrule(lr){2-10}
& \multirow{4}{*}{16} & ICL  & 48.60 & 75.60 & 94.30 & 60.20 & 81.20 & 28.10 & 64.67 \\
&                     & LoRA & 58.80 & 73.50 & 69.70 & 56.40 & 51.20 & 25.00 & 55.77 \\
&                     & LP   & 72.90 & 73.30 & 75.10 & 61.60 & 61.10 & 38.30 & 63.72 \\
&                     & DyVec& 59.90 & 82.90 & 91.30 & 68.20 & 81.80 & 24.60 & \textbf{68.12} \\
\midrule

\multirow{12}{*}{Qwen} 
& \multirow{4}{*}{4} & ICL  & 29.10 & 37.80 & 94.20 & 52.10 & 72.20 & 25.30 & 51.78 \\
&                     & LoRA & 50.40 & 53.80 & 53.30 & 60.80 & 64.70 & 10.90 & 48.98 \\
&                     & LP   & 48.60 & 48.50 & 47.60 & 58.90 & 44.60 & 19.30 & 44.58 \\
&                     & DyVec& 51.70 & 46.80 & 63.00 & 59.90 & 85.00 & 24.10 & \textbf{55.08} \\
\cmidrule(lr){2-10}
& \multirow{4}{*}{8} & ICL  & 34.50 & 56.00 & 94.90 & 56.80 & 75.90 & 26.40 & 57.42 \\
&                    & LoRA & 50.90 & 64.40 & 53.70 & 53.00 & 36.70 & 6.90  & 44.27 \\
&                    & LP   & 51.60 & 51.30 & 54.40 & 60.30 & 59.80 & 24.50 & 50.32 \\
&                    & DyVec& 53.90 & 50.50 & 92.70 & 60.10 & 80.40 & 25.20 & \textbf{60.47} \\
\cmidrule(lr){2-10}
& \multirow{4}{*}{16} & ICL  & 63.40 & 79.60 & 94.20 & 53.30 & 87.90 & 29.30 & \textbf{67.95} \\
&                     & LoRA & 66.70 & 72.60 & 53.70 & 57.70 & 51.30 & 7.80  & 51.63 \\
&                     & LP   & 61.30 & 53.10 & 73.70 & 63.30 & 65.30 & 25.90 & 57.10 \\
&                     & DyVec& 67.00 & 50.50 & 91.00 & 58.70 & 68.50 & 26.20 & 60.32 \\
\midrule

\multirow{12}{*}{DeepSeek} 
& \multirow{4}{*}{4} & ICL  & 39.60 & 47.90 & 92.90 & 53.10 & 70.28 & 10.51 & 52.38 \\
&                     & LoRA & 36.50 & 60.00 & 32.00 & 55.10 & 4.00   & 12.10 & 33.28 \\
&                     & LP   & 57.00 & 56.60 & 51.80 & 53.20 & 62.50 & 36.50 & 52.93 \\
&                     & DyVec& 48.90 & 62.10 & 80.70 & 59.30 & 80.30 & 28.40 & \textbf{59.95} \\
\cmidrule(lr){2-10}
& \multirow{4}{*}{8} & ICL  & 25.20 & 67.70 & 90.50 & 59.20 & 72.50 & 16.30 & 55.23 \\
&                    & LoRA & 40.40 & 42.80 & 54.90 & 58.20 & 34.20 & 21.70 & 42.03 \\
&                    & LP   & 59.80 & 63.50 & 57.10 & 61.20 & 65.90 & 35.80 & 57.22 \\
&                    & DyVec& 49.60 & 50.60 & 92.50 & 64.40 & 78.80 & 35.20 & \textbf{61.85} \\
\cmidrule(lr){2-10}
& \multirow{4}{*}{16} & ICL  & 57.80 & 68.00 & 93.90 & 67.90 & 77.56 & 9.85  & 62.50 \\
&                     & LoRA & 55.20 & 54.30 & 85.10 & 62.20 & 36.30 & 11.00 & 50.68 \\
&                     & LP   & 65.50 & 80.00 & 61.20 & 62.30 & 73.10 & 38.00 & 63.35 \\
&                     & DyVec& 64.30 & 76.00 & 91.50 & 67.90 & 72.70 & 28.90 & \textbf{66.88} \\

\bottomrule
\end{tabular}

\caption{Evaluation results of ICL, LoRA, LP, and DyVec across models (LLaMA, Qwen, DeepSeek), dataset sizes (4/8/16 shots), and six tasks. \textbf{Bold} indicates best performance in each block. The \cellcolor[gray]{0.95}gray column highlights average performance.}
\label{tab:Main Results wo}
\end{table*}

\subsection{Main Results}
\paragraph{Comparison with Adaptation Baselines.}
We evaluate DyVec on six diverse tasks, using the accuracy as the main metric. We report full results in Table~\ref{tab:Main Results wo}, and summarize the averaged performance in Table~\ref{tab:avg performance}.

As shown in Table~\ref{tab:Main Results wo}, DyVec outperforms both few-shot ICL and the LoRA tuning method across most settings. This demonstrates DyVec’s ability to construct effective task representations from limited labeled data, without requiring any parameter updates or additional training data.
Table~\ref{tab:avg performance} further quantifies the relative improvements across three models. In low-resource settings, DyVec achieves 8.5\% relative improvement over 4-shot ICL, 39.5\% over LoRA and 14.3\% over LP; under 8-shot, the gains remain substantial at 8.5\%, 31.5\% and 10.6\%, respectively. These results confirm DyVec’s strong generalization ability in low-resource scenarios.
\begin{table}[H]
\centering
\setlength{\tabcolsep}{1pt}
\renewcommand{\arraystretch}{1.2}
\footnotesize
\begin{tabular}{ccccc}
\toprule
\textbf{DataSize} & \textbf{ICL} & \textbf{LoRA} & \textbf{LP} & \textbf{DyVec} \\
\midrule
4   & 51.17 {\color{red}\scriptsize (↑8.5\%)}  
    & 39.78 {\color{red}\scriptsize (↑39.6\%)}  
    & 48.56 {\color{red}\scriptsize (↑14.3\%)}  
    & 55.51 \\
8   & 56.13 {\color{red}\scriptsize (↑8.6\%)}  
    & 46.31 {\color{red}\scriptsize (↑31.5\%)}  
    & 55.08 {\color{red}\scriptsize (↑10.6\%)}  
    & 60.93 \\
16  & 65.04 {\color{red}\scriptsize (↑0.1\%)}  
    & 52.69 {\color{red}\scriptsize (↑23.6\%)}  
    & 61.39 {\color{red}\scriptsize (↑6.1\%)}  
    & 65.11 \\
\bottomrule
\end{tabular}
\vspace{1mm}
\caption{
Average performance across models and tasks at varying data sizes. Relative improvements of DyVec over each baseline are highlighted in red.
}
\label{tab:avg performance}
\end{table}

\paragraph{Comparison with ICV Methods.}
We compare DyVec with three ICV baselines: Task Vector (TV), Function Vector (FV) and Multimodal Task Vector (MTV), under 8-shot settings using LLaMA-2-7B-Chat, with the accuracy as the primary evaluation metric. FV uses a \textit{Causal Indirect Effect}-based signal which require inferencing on extensive data to locate informative attention head positions to inject vector, while TV use validation set to select injection position for each task.

As shown in Table~\ref{tab:icv_comparison}, DyVec outperforms all ICV methods by large margins, demonstrating the clear advantage of dynamic over other ICV methods.
\begin{table}[H]
\centering
\setlength{\tabcolsep}{4pt}
\renewcommand{\arraystretch}{1.2}
\footnotesize
\begin{tabular}{lccc}
\toprule
\textbf{Model} & \textbf{FV} & \textbf{TV} & \textbf{DyVec} \\
\midrule
gpt-j-6b 
& 42.27 {\color{red}\scriptsize (↑32.1\%)} 
& 37.95 {\color{red}\scriptsize (↑47.0\%)} 
& 55.80 \\
llama-2-7b-chat 
& 45.47 {\color{red}\scriptsize (↑33.0\%)} 
& 22.47 {\color{red}\scriptsize (↑168.9\%)} 
& 60.48 \\
llama-2-13b-chat 
& 43.73 {\color{red}\scriptsize (↑43.2\%)} 
& 40.70 {\color{red}\scriptsize (↑53.9\%)} 
& 62.63 \\
\bottomrule
\end{tabular}
\vspace{1mm}
\caption{
Average accuracy across six tasks. Red text indicates relative improvement of DyVec over each baseline.
}
\label{tab:icv_comparison}
\end{table}

\subsection{Inference Efficiency Analysis}
\begin{figure}[H]
    \centering
    \includegraphics[width=1.0\linewidth]{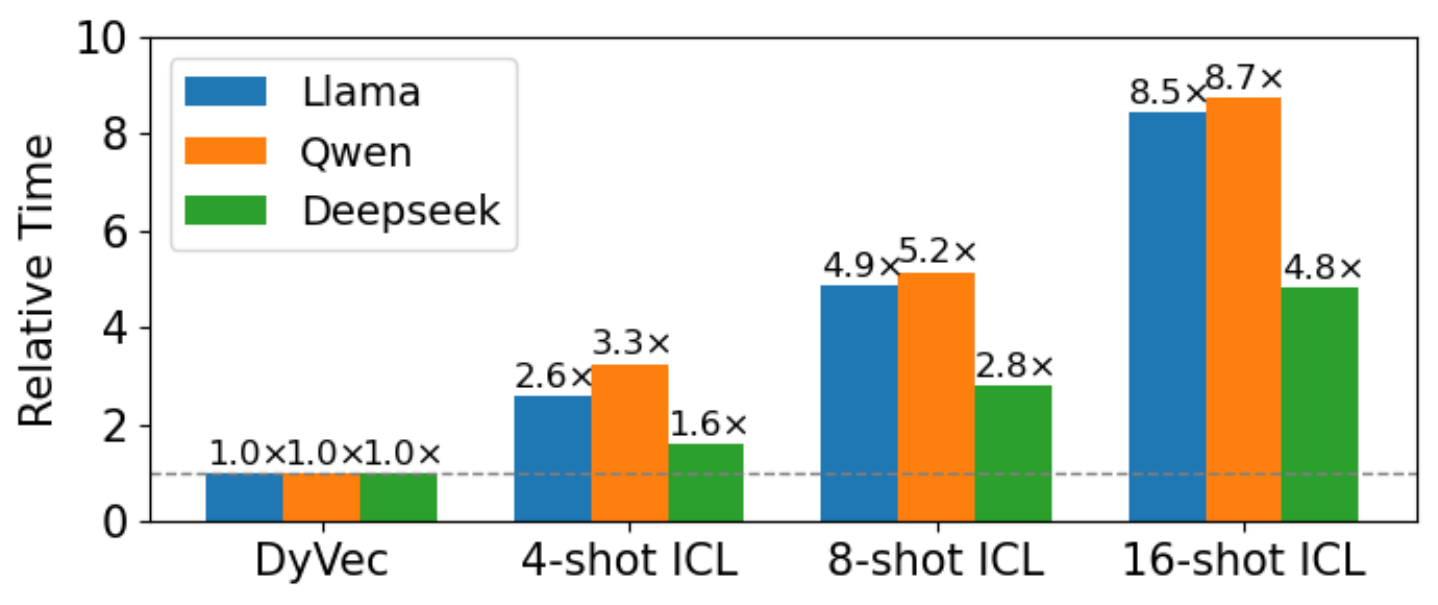}
    \caption{Relative inference time across different models and methods.}
    \label{fig:Time}
\end{figure}

To assess efficiency, we measure the total runtime on an A100 GPU for completing six tasks using DyVec and Few-shot ICL, with 1000 test samples per task. Results are shown in Figure~\ref{fig:Time}.

Unlike ICL, which incurs increasing overhead due to longer prompts, DyVec employs a lightweight inference-time intervention that requires minimal computational cost. Despite its efficiency, DyVec not only matches but significantly outperforms ICL in accuracy, demonstrating that effective task adaptation can be achieved without the expense of costly prompt-based conditioning. Detailed timing statistics are provided in Appendix~\ref{sec:Time}.

\subsection{Ablation Study}

To better understand the impact of key design components in DyVec, we conduct ablation studies under 8-shot settings across three tasks (NHSD, ADE, AG\_News) and three LLMs (LLaMA, Qwen, DeepSeek). 
Specifically, we examine the effects of: (1) the source of representations used for vector construction, (2) the EQR strategy to compute robust representations, (3) the granularity of dynamic latent segment $d_{\mu}$, and (4) the robustness of DyVec to changes in latent segment source.

\paragraph{Semantically Aggregated Latent Representations Enable More Informative Vector Construction.}
To evaluate the impact of representation source on vector construction, we compare two approaches. Prior methods typically use raw \textit{Attention Head Activations} (AHA, Eq.~\ref{eq:attention}) extracted directly from selected attention heads~\cite{huang2024multimodal}. In contrast, DyVec adopts a more structured \textit{Semantically Aggregated Representation} (SAR, Eq.~\ref{eq:proj_output}), obtained from the output projections of the MHA module after cross-head fusion via a learned linear transformation. For fair comparison, the SAR in DyVec is segmented into $S=H$ segments to match the number of attention heads.

As shown in Table~\ref{tab:output_comparison}, SAR consistently outperforms AHA across all three models, highlighting the superior informativeness and generalization of semantically aggregated latent representations as the source of vector construction.
\renewcommand{\arraystretch}{1.0}
\begin{table}[!t]
\centering
\footnotesize
\setlength{\tabcolsep}{1pt}
\begin{tabular}{ccccc|>{\columncolor[gray]{0.9}}c}
\toprule
\textbf{Model} & \textbf{Method} & NHSD & ADE & AG\_News & \textbf{Avg.} \\
\midrule
\multirow{2}{*}{LLaMA-2-7B-Chat} 
 & AHA        & 51.60 & 64.20 & 77.40 & 64.40 \\
 & SAR  & 49.00 & 67.40 & 79.20 & \textbf{65.20} \\
\midrule
\multirow{2}{*}{LLaMA-2-13B-Chat} 
 & AHA        & 50.10 & 54.50 & 82.10 & 62.23 \\
 & SAR  & 52.90 & 66.60 & 86.00 & \textbf{68.50} \\
\midrule
\multirow{2}{*}{Qwen2-7B} 
 & AHA        & 51.70 & 55.70 & 78.90 & 62.10 \\
 & SAR  & 53.90 & 60.10 & 80.40 & \textbf{64.80} \\
\midrule
\multirow{2}{*}{DeepSeek-7B-LLM} 
 & AHA        & 47.20 & 64.80 & 77.20 & 63.07 \\
 & SAR  & 49.60 & 64.40 & 78.80 & \textbf{64.27} \\
\midrule
\multirow{2}{*}{GPT-J-6B} 
 & AHA        & 53.20 & 50.20 & 69.50 & 57.63 \\
 & SAR  & 56.10 & 59.20 & 68.50 & \textbf{61.27} \\
\bottomrule
\end{tabular}
\caption{Comparison of sources for vector construction.}
\label{tab:output_comparison}
\end{table}



\begin{figure}[!t]
    \centering
    \includegraphics[width=1.0\linewidth]{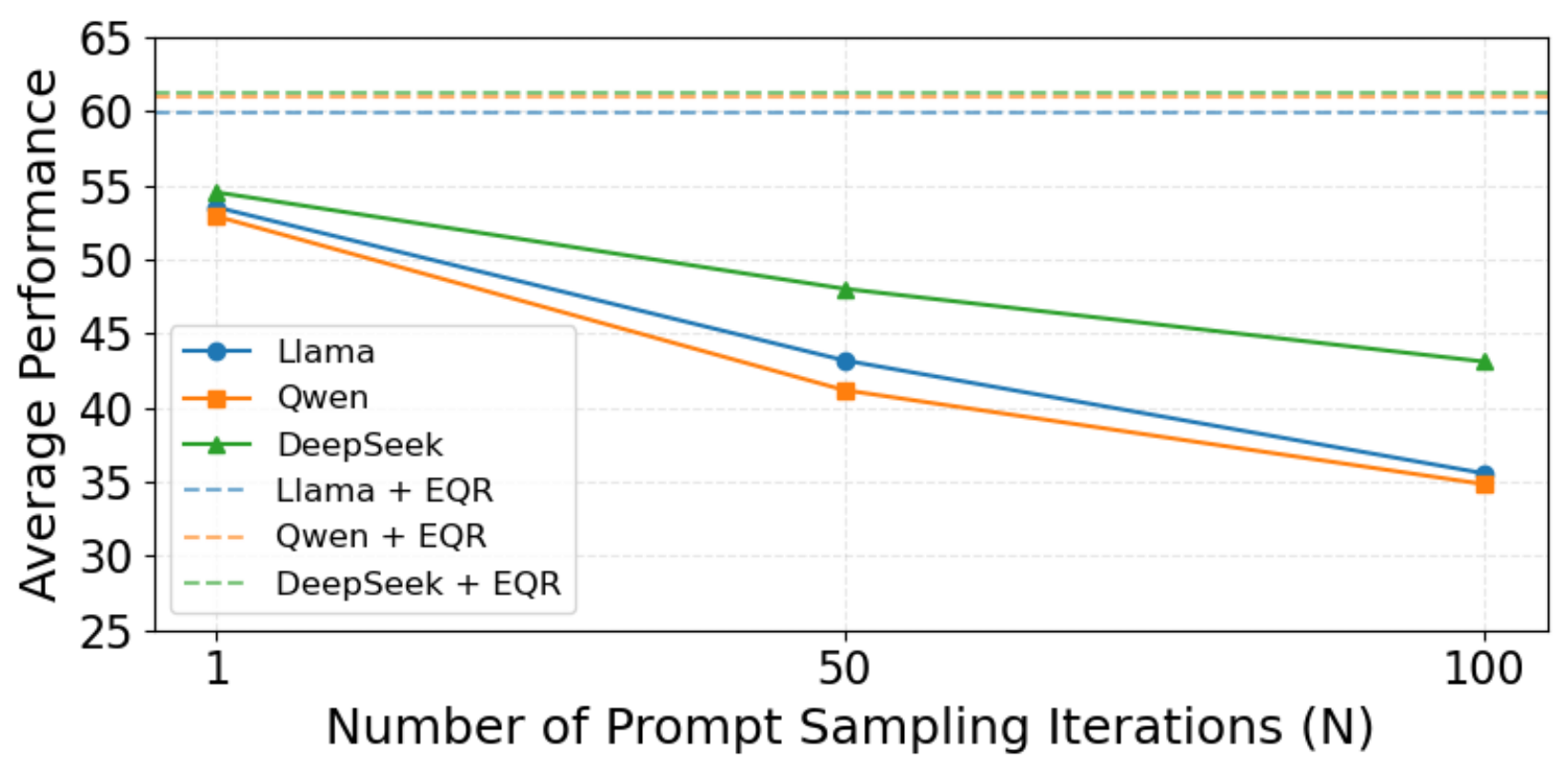}
    \caption{Effectiveness of the EQR strategy in DyVec across different models. Results are averaged over three tasks with 8-shot data. Solid lines represent performance using different numbers of randomly constructed prompts ($N = 1, 50, 100$), while dashed lines indicate performance using  EQR strategy.}
    \label{fig:N}
\end{figure}
\renewcommand{\arraystretch}{1.1}
\begin{table}[h]
\centering
\footnotesize
\setlength{\tabcolsep}{2pt}
\begin{tabular}{lcccccc|>{\columncolor[gray]{0.9}}c}
\toprule
\textbf{Method} & NHSD & Sar. & SST2 & ADE & AG. & TREC6 & \textbf{Avg.} \\
\midrule
TV           & 25.21  & 26.81 & 27.82  & 23.51  & 18.34  & 13.11  & 22.47 \\
+ EQR        & 34.65  & 29.53 & 32.76  & 34.09  & 26.42  & 14.95  & \textbf{28.74} \\
\midrule
DyVec$^\prime$ & 46.70  & 58.70 & 77.50  & 58.20  & 58.90  & 34.30  & 55.72 \\
+ EQR        & 49.00  & 50.40 & 91.50  & 67.40  & 79.20  & 25.40  & \textbf{60.48} \\
\bottomrule
\end{tabular}
\caption{Effectiveness of the EQR strategy on different ICV methods. DyVec$^\prime$ denotes DyVec without EQR.}
\label{tab:ablation_EQR}
\end{table}

\begin{table}[ht]
\centering
\footnotesize
\setlength{\tabcolsep}{4pt}
\begin{tabular}{llccc|>{\columncolor[gray]{0.9}}c}
\toprule
\textbf{Model} & \textbf{Method} & NHSD & ADE & Ag\_news & \textbf{Avg.} \\
\midrule
\multirow{3}{*}{LLaMa} 
    & Full     & 44.30 & 55.60 & 60.80 & 53.57 \\
    & Partial  & 50.60 & 53.40 & 60.20 & 54.73 \\
    & EQR                 & 48.50 & 56.90 & 64.20 & \textbf{56.53} \\
\midrule
\multirow{3}{*}{Qwen} 
    & Full     & 52.80 & 54.10 & 73.60 & 60.17 \\
    & Partial  & 50.60 & 55.00 & 81.00 & 62.20 \\
    & EQR                 & 51.70 & 59.90 & 85.00 & \textbf{65.53} \\
\midrule
\multirow{3}{*}{DeepSeek} 
    & Full     & 47.40 & 51.60 & 71.30 & 56.77 \\
    & Partial  & 50.60 & 50.00 & 77.90 & 59.50 \\
    & EQR                 & 48.90 & 59.30 & 80.30 & \textbf{62.83} \\
\bottomrule
\end{tabular}
\caption{Ablation Analysis of EQR with Respect to Permutation Variants.}
\label{tab:EQR-results}
\end{table}

\renewcommand{\arraystretch}{1.0}
\begin{table}[!t]
\centering
\footnotesize
\setlength{\tabcolsep}{5pt}
\begin{tabular}{ccccc|>{\columncolor[gray]{0.9}}c}
\toprule
\textbf{Model} & S & NHSD & ADE & AG\_News & \textbf{Avg.} \\
\midrule
\multirow{4}{*}{LLaMA} 
 & $1$     & 54.60 & 53.50 & 71.40 & 59.83 \\
 & $H/2$   & 54.60 & 61.10 & 74.30 & 63.33 \\
 & $H$     & 49.00 & 67.40 & 79.20 & 65.20 \\
 & $2H$    & 53.10 & 66.00 & 78.40 & \textbf{65.83} \\
\midrule
\multirow{4}{*}{Qwen} 
 & $1$     & 50.60 & 51.20 & 28.70 & 43.50 \\
 & $H/2$   & 53.10 & 59.70 & 79.90 & 64.23 \\
 & $H$     & 53.90 & 60.10 & 80.40 & 64.80 \\
 & $2H$    & 57.70 & 63.40 & 83.70 & \textbf{68.27} \\
\midrule
\multirow{4}{*}{DeepSeek} 
 & $1$     & 50.60 & 50.00 & 68.40 & 56.33 \\
 & $H/2$   & 55.30 & 61.10 & 71.30 & 62.57 \\
 & $H$     & 49.60 & 64.40 & 78.80 & 64.27 \\
 & $2H$    & 49.40 & 69.10 & 77.00 & \textbf{65.17} \\
\bottomrule
\end{tabular}
\caption{
Effect of latent segmentation granularity ($S$) on DyVec performance across models and tasks.
}
\label{tab:Latent}
\end{table}

\paragraph{EQR Enhances Representation Robustness Across Methods.}

DyVec incorporates an \textit{Exhaustive Query Rotation} (EQR) strategy, which systematically rotates the query position within a fixed demonstration set and averages the resulting representations to compute a robust task representation for vector construction. To evaluate its effectiveness, we compare EQR against a baseline that constructs prompts via random shuffling and computes the representation by averaging outputs over $N$ iterations.

As shown in Figure~\ref{fig:N}, EQR consistently achieves the highest average performance across all models. Importantly, EQR is orthogonal to other ICV-based representation strategies and can be seamlessly integrated as a complementary enhancement for robust vector construction, as further evidenced in Table~\ref{tab:ablation_EQR}.

To further analyze EQR, we compare it against both partial and full permutation baselines, which randomly shuffle demonstrations either partially or entirely. As shown in Table~\ref{tab:EQR-results}, EQR consistently yields more robust representations across tasks and models, highlighting its effectiveness over naive permutation strategies.



\paragraph{Finer Latent Segmentation Leads to Better Vectors.}

In DyVec, the semantically aggregated latent representation $\overline{\mathbf{o}}_{(i)}$ is divided into $S$ segments, each corresponding to an independently optimized subspace. The default setting $S = H$ (number of attention heads) serves as a standard granularity. We compare three variants:
(1) \textit{Fine-grained}: $S = 2H$;
(2) \textit{Coarse-grained}: $S = H/2$;
(3) \textit{Layer-grained}: $S = 1$.
As shown in Table~\ref{tab:Latent}, finer segmentation ($S = 2H$) consistently yields the best performance. Coarser settings lead to noticeable drops, with $S = 1$ reducing DyVec to a global layer-shaped representation similar to the Task Vector baseline. This confirms that finer-grained intervention captures richer task signals, while overly coarse representations lack sufficient capacity.


\begin{table*}[!t]
\centering
\small
\begin{tabular}{lccc|c}
\toprule
\textbf{Method} & \textbf{Vector Construction} & \textbf{Search/Opt.} & \textbf{Inference} & \textbf{Total Cost} \\
\midrule
FV & 2 forward passes ($<$1s) & CIE ($\sim$59min) + Layer search (114s $\times$ 32 $\approx$ 61min) & 114s & $\sim$120min \\
TV & 1 forward pass ($<$1s) & Layer search (118s $\times$ 32 $\approx$ 63min) & 118s & $\sim$63min \\
DyVec (Ours) & $N$ forward passes ($\sim$2s) & REINFORCE ($\sim$17min) & 120s & $\sim$19min \\
\bottomrule
\end{tabular}
\caption{Efficiency comparison across methods. DyVec achieves significant cost reduction compared to FV and TV.}
\label{tab:efficiency}
\end{table*}

\paragraph{Robustness of Task Injection under Latent Segment Replacement.}
To further assess the robustness and generalization ability of DyVec, we conduct a supplementary experiment by decoupling the sources of vector construction and intervention configuration. Specifically, we reconstruct the latent segments $u_{(i,j)}^{\mathcal{D}_X}$ from a dataset $\mathcal{D}_X$, while keeping the selected intervention positions $\mathcal{Y}^{*\mathcal{D}_K}$ and injection function $\mathcal{L}^{\mathcal{D}_K}$ fixed, as determined on a separate dataset $\mathcal{D}_K$. Formally,
\begin{equation}
\theta^{X \leftarrow K} = \{ u_{(i,j)}^{\mathcal{D}_X} \mid (i,j) \in \mathcal{Y}^{*\mathcal{D}_K} \}
\label{eq:transfer}
\end{equation}
In this setting, only the latent segments $u_{(i,j)}$ are updated using new examples from $\mathcal{D}_X$, while the injection location and strategy remain fixed. This ensures that the intervention mechanism is consistent, isolating the effect of changing the latent source.

Despite being derived from entirely different examples, the injected vectors still lead to strong performance across multiple tasks—sometimes even surpassing the original configuration. These results highlight DyVec’s robustness to variation in vector sources. Detailed results are provided in Appendix~\ref{sec:Task Representation Transferability}.

\subsection{Efficiency Analysis}
We have profiled the vector construction and training time of \textbf{DyVec} and the key ICV baselines (\textbf{FV}, \textbf{TV}).  
The experiment was conducted on the \textit{AG\_News} task (data size $N=8$) using the \textit{LLaMA-2-7B-Chat} model, running on a single \textit{NVIDIA A40 GPU}.  
The detailed breakdown of the time cost for each method is presented in Table~\ref{tab:efficiency}. Results show that DyVec substantially reduces computational cost: compared with FV (which requires costly contrastive instance extraction and layer search) and TV (which also depends on exhaustive layer search), DyVec achieves a much lower overall cost by leveraging efficient optimization.

\section{Conclusion}

We propose DyVec, a novel ICV method for efficient and robust inference-time task adaptation in LLMs. DyVec addresses key limitations of prior approaches through three innovations: (1) extracting semantically aggregated latent representations as the source for vector construction, (2) employing EQR to compute robust task representations, and (3) performing dynamic latent segmentation and flexible vector injection via REINFORCE optimization.
Experiments across diverse tasks and model scales show that DyVec consistently outperforms few-shot ICL, LoRA, and previous ICV baselines, while preserving the efficiency of zero-shot inference. Beyond strong performance, DyVec offers a lightweight, generalizable framework for vector-based intervention, deepening our understanding of latent task representations in LLMs.




\section*{Limitations}

While DyVec demonstrates promising results on classification and lexical generation tasks, it still has several limitations. First, the current evaluation is primarily focused on classification tasks, with a limited number of generation tasks included for auxiliary analysis. Further validation on more diverse and complex tasks, such as multi-hop reasoning, dialogue, or instruction following, is needed to assess broader applicability. Second, our intervention strategy is designed manually and remains fixed during inference. Although we experiment with different positions and segmentations, the current approach does not explore adaptive or learned intervention mechanisms, which may further enhance performance or stability.

\section*{Ethics Statement}
\paragraph{Use of AI Assistants}
We have employed ChatGPT as a writing assistant, primarily for polishing the text after the initial composition.

\section*{Acknowledgments}
This work is supported by the National Natural Science Foundation of China (62076008).

\bibliographystyle{acl_natbib}

\clearpage

\appendix
\section{Prompt Construction Details}
\label{sec:prompt construction}
For all tasks in our experiments, we adopt a unified prompt format based on a question–answering template. Each example is framed as a pair of input–output sentences in the form of:
\begin{quote}
\texttt{Q: [input text] \textbackslash n A: [label]}
\end{quote}

Few-shot ICL prompts are constructed by concatenating $k$ such (Q, A) pairs as demonstrations, followed by a query in the same format but without the answer. A special delimiter token \texttt{<|endoftext|>} is prepended to mark the start of the prompt. All tasks share the same prompt structure, with only the task-specific examples substituted.

\vspace{1mm}
\noindent\textbf{Example (SST2, 4-shot):}
\begin{quote}
\small
\texttt{<|endoftext|>}  
\texttt{Q: plenty of warmth to go around , with music and laughter and the love of family \textbackslash n A: positive}

\texttt{Q: comfort \textbackslash n A: positive}

\texttt{Q: lacks a strong narrative \textbackslash n A: negative}

\texttt{Q: attal 's hang-ups surrounding infidelity are so old-fashioned \textbackslash n A: negative}

\texttt{Q: painful elegy \textbackslash n A:}
\end{quote}

\vspace{1mm}
\noindent\textbf{Example (SST2, 0-shot):}
\begin{quote}
\small
\texttt{<|endoftext|> Q: painful elegy \textbackslash n A:}
\end{quote}

This template is applied uniformly across all classification and generation tasks, enabling a fair comparison between few-shot and intervention-based methods.

\section{Data Construction Details}
\label{sec:Data Construction Details}

For all classification tasks used in our experiments, we construct training data with an emphasis on \textit{diversity}, aiming to help in-context learning (ICL) better capture task semantics and label structure. When selecting $k$ labeled examples, we ensure that different label categories are represented as evenly as possible.

For example, the \textbf{AG\_News} dataset contains four categories: \textit{World}, \textit{Business}, \textit{Sports}, and \textit{Science}. When $k=4$, we select one example from each class. Below is a subset used under this setting:

\begin{quote}
\textbf{Input:} Linux puts another financial feather in its cap...
\textbf{Output:} Science

\textbf{Input:} War in Iraq Did Not Make World Safer, Annan Says...
\textbf{Output:} World

\textbf{Input:} US Treasuries cut early gains on jobless drop...
\textbf{Output:} Business

\textbf{Input:} Closing Ceremonies Host city Athens bid a final farewell...
\textbf{Output:} Sports
\end{quote}

We apply the same strategy to other datasets: for instance, in \textbf{SST2} (a binary sentiment task), we use two positive and two negative sample when $k=4$; in \textbf{TREC6}, we attempt to sample across all six question categories. This ensures that each in-context demonstration set provides broad task coverage, which is especially important in the few-shot setting.

\section{Baselines}
\label{sec:baselines}
\paragraph{Few-Shot In-Context Learning (ICL)}
The model is presented with $4/8/16$ labeled examples directly within the input prompt, without any modification to its parameters. This non-parametric adaptation method is widely used due to its simplicity and general applicability.

\paragraph{LoRA Fine-tuning}
This parameter-efficient fine-tuning approach introduces low-rank adapters into the attention projection layers. By updating only a small number of additional parameters while keeping the base model weights frozen, LoRA enables efficient adaptation to specific tasks. Implementation details for LoRA are provided in Appendix~\ref{sec:LoRA}.

\paragraph{Task Vector (TV)}
This method encodes $k$ demonstrations combined with a dummy query and extracts the representation of the last token from an intermediate layer as the ICV. During inference, this vector replaces the corresponding last token representation in the same layer. We evaluate TV at different layers and report the performance of the best-performing layer based on test set results.

\paragraph{Function Vector (FV)}
The ICV is derived by averaging the outputs of key attention heads over a small validation subset. This vector is then added to the last token representation at a selected layer during inference, modulating the model's behavior for the target task. Similar to TV, we evaluate FV across multiple layers and report the test performance of the best-performing configuration.

\paragraph{Linear Probing (LP)}
The core idea of LP is to freeze the pretrained LLM and train a lightweight classifier on top of its final hidden representations before the output logits layer. This provides a simple and efficient adaptation baseline, and we report its performance under the same settings as other baselines.
\renewcommand{\arraystretch}{1.0}
\begin{table}[H]
\centering
\footnotesize
\begin{tabular}{c@{\hspace{12pt}}l@{\hspace{12pt}}c}
\toprule
\textbf{Model} & \textbf{Method} & \textbf{Time $\downarrow$ (min)} \\
\midrule
\multirow{5}{*}{\textbf{LLaMA}} 
  & LoRA     & 5.16 \\
  & DyVec    & 6.12 \\
  & 4-shot   & 15.75 \\
  & 8-shot   & 29.90 \\
  & 16-shot  & 51.73 \\
\midrule
\multirow{5}{*}{\textbf{Qwen}} 
  & LoRA     & 4.12 \\
  & DyVec    & 4.95 \\
  & 4-shot   & 16.16 \\
  & 8-shot   & 25.51 \\
  & 16-shot  & 43.29 \\
\midrule
\multirow{5}{*}{\textbf{DeepSeek}} 
  & LoRA     & 6.50 \\
  & DyVec    & 9.40 \\
  & 4-shot   & 15.15 \\
  & 8-shot   & 26.41 \\
  & 16-shot  & 45.51 \\
\bottomrule
\end{tabular}
\caption{Inference time (in minutes) across different models and methods.}
\label{tab:Time}
\end{table}

\begin{table*}[!t]
\centering
\footnotesize
\setlength{\tabcolsep}{6pt} 
\renewcommand{\arraystretch}{1.0} 
\begin{tabular}{cc|cccccc|c}
\toprule
\textbf{Model} & \textbf{Method} & \textbf{NHSD} & \textbf{Sarcasm} & \textbf{SST2} & \textbf{ADE} & \textbf{AG\_News} & \textbf{TREC6} & \textbf{Avg} \\
\midrule
\multirow{4}{*}{llama} 
 & EQR     & 49.00 & 50.40 & 91.50 & 67.40 & 79.20 & 25.40 & \textbf{60.48} \\
 & N = 1   & 46.70 & 58.70 & 77.50 & 58.20 & 58.90 & 34.30 & 55.72 \\
 & N = 50  & 49.10 & 50.10 & 57.10 & 54.40 & 57.70 & 24.10 & 48.75 \\
 & N = 100 & 50.60 & 50.50 & 55.10 & 50.00 & 25.10 & 24.10 & 42.57 \\
\midrule
\multirow{4}{*}{Qwen} 
 & EQR     & 34.50 & 56.00 & 94.90 & 56.80 & 75.90 & 26.40 & \textbf{57.42} \\
 & N = 1   & 42.70 & 43.40 & 85.70 & 52.40 & 26.10 & 17.70 & 44.67 \\
 & N = 50  & 49.40 & 50.50 & 66.30 & 50.40 & 36.40 & 24.10 & 46.18 \\
 & N = 100 & 50.60 & 50.50 & 46.30 & 50.20 & 25.10 & 16.50 & 39.87 \\
\midrule
\multirow{4}{*}{deepseek} 
 & EQR     & 25.20 & 67.70 & 90.50 & 59.20 & 72.50 & 16.30 & \textbf{55.23} \\
 & N = 1   & 45.80 & 60.20 & 86.50 & 63.00 & 52.10 & 33.20 & 56.80 \\
 & N = 50  & 48.00 & 50.50 & 65.90 & 47.60 & 68.40 & 27.30 & 51.28 \\
 & N = 100 & 42.00 & 50.50 & 53.70 & 50.00 & 29.90 & 24.20 & 41.72 \\
\bottomrule
\end{tabular}
\caption{Performance comparison of different models using the EQR method and varying the number of prompt
sampling iterations ($N$) across six datasets.}
\label{tab:prompt}
\end{table*}

\section{Inference Time}
\label{sec:Time}

To further understand the practical efficiency of DyVec, we compare the inference time of different adaptation methods across LLaMA, Qwen, and DeepSeek models. All evaluations are conducted on the same hardware with a fixed test set size of 1000 examples per task, ensuring a fair comparison of decoding speed.
Results are reported in Table~\ref{tab:Time}.
We observe that:

Few-shot ICL incurs the highest inference cost, and this cost scales roughly linearly with the number of in-context examples. For instance, going from 4-shot to 16-shot increases decoding time by over 3 times for all models. This highlights the inefficiency of prompt-based adaptation at scale.
LoRA offers the fastest inference, since it only relies on a fixed set of fine-tuned parameters and incurs no additional prompt-related overhead.
DyVec achieves a favorable balance: while slightly slower than LoRA due to runtime vector injection, it consistently outperforms ICL in efficiency, especially in higher-shot settings. 

These results demonstrate that DyVec retains non-parametric generality without the latency penalty of few-shot prompting, making it more suitable for efficient inference scenarios.
\renewcommand{\arraystretch}{1.0}
\begin{table*}[!t]
\centering
\footnotesize
\setlength{\tabcolsep}{4pt} 

\begin{tabular}{cccllllll}
\toprule
Model & Data size & Method & Antonym & Capitalize & Country-capital & English-french & Present-past & Singular-plural \\
\midrule
\multirow{9}{*}{Llama} & \multirow{3}{*}{4} 
& ICL & \textbf{67.66} & \textbf{100.00} & \textbf{97.62} & \textbf{78.57} & 95.24 & \textbf{100.00} \\
& & LoRA & 41.67 & 56.47 & 42.86 & 21.78 & 60.66 & 93.02 \\
& & DyVec & 56.55 & 98.24 & \textbf{97.62} & 66.36 & \textbf{96.72} & \textbf{100.00} \\
\cmidrule{2-9}
& \multirow{3}{*}{8}
& ICL & \textbf{69.05} & \textbf{100.00} & \textbf{97.62} & \textbf{85.71} & \textbf{100.00} & \textbf{100.00} \\
& & LoRA & 40.67 & 74.71 & 88.10 & 31.21 & 88.52 & 90.70 \\
& & DyVec & 49.80 & \textbf{100.00} & \textbf{97.62} & 72.75 & \textbf{100.00} & \textbf{100.00} \\
\cmidrule{2-9}
& \multirow{3}{*}{16}
& ICL & \textbf{70.24} & \textbf{100.00} & \textbf{100.00} & \textbf{80.95} & \textbf{100.00} & \textbf{100.00} \\
& & LoRA & 53.17 & 85.29 & 88.10 & 41.54 & 90.16 & 97.67 \\
& & DyVec & 63.10 & \textbf{100.00} & 95.24 & 78.32 & \textbf{100.00} & \textbf{100.00} \\
\midrule
\multirow{9}{*}{Qwen} & \multirow{3}{*}{4}
& ICL & \textbf{67.86} & \textbf{100.00} & 95.24 & \textbf{85.71} & 97.62 & 97.62 \\
& & LoRA & 15.87 & 53.53 & 35.71 & 10.84 & 31.15 & 34.88 \\
& & DyVec & 53.17 & \textbf{100.00} & \textbf{97.62} & 76.90 & \textbf{100.00} & \textbf{100.00} \\
\cmidrule{2-9}
& \multirow{3}{*}{8}
& ICL & \textbf{69.44} & \textbf{100.00} & 95.24 & \textbf{85.71} & 97.62 & 97.62 \\
& & LoRA & 35.12 & 54.71 & 47.62 & 32.12 & 54.10 & 76.74 \\
& & DyVec & 55.36 & \textbf{100.00} & \textbf{95.24} & 75.38 & \textbf{100.00} & \textbf{100.00} \\
\cmidrule{2-9}
& \multirow{3}{*}{16}
& ICL & \textbf{72.42} & \textbf{100.00} & 95.24 & \textbf{85.71} & 97.62 & 97.62 \\
& & LoRA & 52.98 & 74.71 & 78.57 & 50.35 & 83.61 & 83.72 \\
& & DyVec & 61.11 & \textbf{100.00} & \textbf{100.00} & 76.39 & \textbf{100.00} & \textbf{100.00} \\
\midrule
\multirow{9}{*}{deepseek} & \multirow{3}{*}{4}
& ICL & \textbf{70.04} & \textbf{97.06} & \textbf{90.48} & \textbf{83.33} & 95.24 & 97.62 \\
& & LoRA & 22.82 & 50.59 & 16.67 & 33.43 & 60.66 & 25.58 \\
& & DyVec & 39.29 & 79.41 & 88.10 & 69.10 & \textbf{96.72} & \textbf{97.67} \\
\cmidrule{2-9}
& \multirow{3}{*}{8}
& ICL & \textbf{70.44} & \textbf{100.00} & 88.10 & \textbf{73.81} & 97.62 & 97.62 \\
& & LoRA & 22.62 & 65.88 & 35.71 & 38.20 & 67.21 & 65.12 \\
& & DyVec & 53.77 & \textbf{100.00} & \textbf{88.10} & 72.64 & \textbf{100.00} & \textbf{100.00} \\
\cmidrule{2-9}
& \multirow{3}{*}{16}
& ICL & \textbf{69.84} & \textbf{100.00} & 90.48 & \textbf{76.19} & 90.48 & 97.62 \\
& & LoRA & 33.33 & 92.94 & 73.81 & 50.96 & 83.61 & 83.72 \\
& & DyVec & 60.32 & \textbf{100.00} & \textbf{92.86} & 73.66 & \textbf{100.00} & \textbf{97.67} \\
\bottomrule
\end{tabular}
\vspace{1mm}
\caption{Accuracy comparison (\%) on six linguistic transformation tasks under different adaptation methods across three models and varying data sizes. Bold numbers indicate the best results in each setting. DyVec consistently achieves competitive or superior performance.}
\label{tab:linguistic task}
\end{table*}
\begin{table*}[!t]
\centering
\footnotesize
\setlength{\tabcolsep}{6pt}
\renewcommand{\arraystretch}{1.0}

\begin{tabular}{c c c c c c c c c}
\toprule
\textbf{Model} & \textbf{$\mathcal{D}_Y$} & \textbf{$\mathcal{D}_X$} & \textbf{NHSD} & \textbf{Sarcasm} & \textbf{SST2} & \textbf{ADE} & \textbf{AG\_News} & \textbf{TREC6} \\
\midrule
\multirow{9}{*}{Llama} 
 & \multirow{3}{*}{4} 
 & 4  & 48.50 & \textbf{57.40} & 55.60 & 56.90 & 64.20 & 26.40 \\
 &  & 8  & \textbf{50.60} & 52.10 & \textbf{90.10} & \textbf{57.50} & \textbf{65.00} & \textbf{27.00} \\
 &  & 16 & \textbf{50.60} & 56.90 & 59.70 & 50.00 & 60.60 & 24.60 \\
 \cmidrule{2-9}
 & \multirow{3}{*}{8} 
 & 4  & \textbf{57.40} & \textbf{50.50} & 73.40 & \textbf{67.60} & 65.00 & 24.10 \\
 &  & 8  & 49.00 & 50.40 & \textbf{91.50} & 67.40 & \textbf{79.20} & 25.40 \\
 &  & 16 & 54.20 & \textbf{50.50} & 84.80 & 50.20 & 76.40 & \textbf{26.20} \\
 \cmidrule{2-9}
 & \multirow{3}{*}{16} 
 & 4  & 54.10 & 65.60 & 82.20 & 51.50 & 56.40 & \textbf{28.70} \\
 &  & 8  & 50.70 & 74.90 & \textbf{92.10} & 50.50 & 78.90 & 27.20 \\
 &  & 16 & \textbf{59.90} & \textbf{82.90} & 91.30 & \textbf{68.20} & \textbf{81.80} & 24.60 \\
\midrule
\multirow{9}{*}{Qwen} 
 & \multirow{3}{*}{4} 
 & 4  & \textbf{51.70} & 46.80 & 63.00 & \textbf{59.90} & \textbf{85.00} & 24.10 \\
 &  & 8  & 50.30 & 50.10 & \textbf{79.80} & 59.00 & 75.30 & \textbf{27.60} \\
 &  & 16 & 50.40 & \textbf{50.50} & 71.70 & 56.10 & 80.80 & 23.20 \\
 \cmidrule{2-9}
 & \multirow{3}{*}{8} 
 & 4  & 50.60 & \textbf{50.50} & 88.00 & \textbf{60.90} & 77.40 & \textbf{25.70} \\
 &  & 8  & 53.90 & \textbf{50.50} & \textbf{92.70} & 60.10 & \textbf{80.40} & 25.20 \\
 &  & 16 & \textbf{59.30} & \textbf{50.50} & 78.00 & 56.10 & 75.50 & 24.40 \\
 \cmidrule{2-9}
 & \multirow{3}{*}{16} 
 & 4  & 50.20 & \textbf{51.10} & \textbf{94.00} & 57.50 & \textbf{73.90} & \textbf{33.80} \\
 &  & 8  & 63.10 & 50.10 & 87.50 & \textbf{59.20} & 50.20 & 29.60 \\
 &  & 16 & \textbf{67.00} & 50.50 & 91.00 & 58.70 & 68.50 & 26.20 \\
\midrule
\multirow{9}{*}{deepseek} 
 & \multirow{3}{*}{4} 
 & 4  & 48.90 & 62.10 & 80.70 & \textbf{59.30} & \textbf{80.30} & 28.40 \\
 &  & 8  & \textbf{50.60} & \textbf{63.70} & \textbf{91.20} & 56.30 & 74.00 & \textbf{41.60} \\
 &  & 16 & \textbf{50.60} & 58.80 & 88.50 & 50.00 & 61.60 & 26.80 \\
 \cmidrule{2-9}
 & \multirow{3}{*}{8} 
 & 4  & \textbf{50.60} & 50.50 & 84.30 & \textbf{65.80} & 62.80 & 24.10 \\
 &  & 8  & 49.60 & \textbf{50.60} & \textbf{92.50} & 64.40 & \textbf{78.80} & \textbf{35.20} \\
 &  & 16 & \textbf{50.60} & 50.50 & 84.50 & 50.10 & 66.70 & 28.30 \\
 \cmidrule{2-9}
 & \multirow{3}{*}{16} 
 & 4  & 50.60 & 69.10 & 86.40 & 50.30 & 38.30 & 23.20 \\
 &  & 8  & 56.10 & \textbf{84.90} & 89.80 & 54.70 & 69.40 & 24.20 \\
 &  & 16 & \textbf{64.30} & 76.00 & \textbf{91.50} & \textbf{67.90} & \textbf{72.70} & \textbf{28.90} \\
\bottomrule
\end{tabular}

\caption{
Dynamic Latent Segments extracted from the dataset $\mathcal{D}_X$, and the intervention positions learned on another dataset ${\mathcal{D}_Y}$. For the same task, fixed intervention positions generalize well across different activation vectors.
}
\label{tab:transfer}
\end{table*}

\begin{table*}[!t]
\centering
\footnotesize
\setlength{\tabcolsep}{6pt}
\renewcommand{\arraystretch}{1.0}
\begin{tabular}{c|c|cccccc}
\toprule
\textbf{Model} & \textbf{Data size} & \textbf{NHSD} & \textbf{Sarcasm} & \textbf{SST2} & \textbf{ADE} & \textbf{AG\_News} & \textbf{TREC6} \\
\midrule
\multirow{3}{*}{LLaMA} 
& 4  & (1,4) & (1,4) & (1,1) & (0,2) & (0,2) & (1,2) \\
& 8  & (1,4) & (1,4) & (1,1) & (0,2) & (1,2) & (1,1) \\
& 16 & (0,1) & (0,1) & (0,4) & (0,1) & (0,1) & (0,1) \\
\midrule
\multirow{3}{*}{Qwen} 
& 4  & (0,1) & (0,4) & (1,1) & (0,2) & (0,4) & (0,1) \\
& 8  & (0,1) & (0,1) & (1,4) & (0,1) & (1,4) & (1,1) \\
& 16 & (0,1) & (1,4) & (1,2) & (0,1) & (0,4) & (0,1) \\
\midrule
\multirow{3}{*}{Deepseek} 
& 4  & (1,4) & (0,1) & (1,1) & (0,1) & (1,4) & (1,1) \\
& 8  & (0,4) & (0,4) & (1,2) & (1,2) & (1,4) & (0,1) \\
& 16 & (0,1) & (0,1) & (0,2) & (0,1) & (0,1) & (0,2) \\
\bottomrule
\end{tabular}
\caption{Optimal intervention configurations $(\alpha, \beta)$ across models and tasks.The number of segmented positions $S$ equals the number of attention heads $H$}
\label{tab:Method}
\end{table*}

\section{LoRA Fine-tuning Details}
\label{sec:LoRA}
To establish a strong parameter-efficient fine-tuning (PEFT) baseline, we employ Low-Rank Adaptation (LoRA) on top of pretrained LLMs. In this section, we provide the detailed configuration and training setup used in our experiments.

We adopt an enhanced LoRA configuration tailored for causal language modeling (CLM). Specifically, we apply LoRA modules to all attention projections: `q\_proj`, `k\_proj`, `v\_proj`, and `o\_proj`, enabling full adaptation within the self-attention mechanism. The rank of the low-rank matrices is denoted by $r$, and we set the scaling factor to $\alpha = 2r$. A dropout rate of 0.2 is used within LoRA to improve training stability. We freeze all original model parameters and fine-tune only the LoRA modules, while explicitly saving the embedding (`embed\_tokens`) and output head (`lm\_head`) layers to ensure correct downstream decoding.

For data preprocessing, we format task-specific input-output pairs and tokenize them using the model’s tokenizer. Dynamic padding is applied to ensure efficient GPU utilization with a padding multiple of 8.

We train the LoRA-augmented model using the HuggingFace `Trainer` API with the following settings:

Batching: Per-device batch size of 4, with gradient accumulation to simulate larger batch sizes.
Optimization: AdamW optimizer with $\beta_1 = 0.9$, $\beta_2 = 0.98$, and weight decay of 0.001 for better convergence. The learning rate follows a cosine scheduler with 20\% warmup.

Precision: Training is performed using bfloat16 (if supported) or fallback to fp16.

Stabilization: Gradient clipping is applied with a norm threshold of 1.0, and gradient checkpointing is enabled to reduce memory usage.

Epochs: The number of training epochs is task-specific, selected via grid search on the development set.

\section{Effect of Exhaustive Query Rotation strategy}

\label{sec:EQR}
In this appendix, we provide a detailed analysis of the Exhaustive Query Rotation (EQR) strategy and its impact on model performance. EQR is designed to enhance the robustness of in-context learning by systematically rotating the query position within the prompt. This approach enables the model to better generalize across different query placements, reducing potential positional biases.

As shown in Table~\ref{tab:prompt}, EQR consistently improves performance across multiple datasets and model architectures compared to Single Prompt and Random Shuffle Averaging. The strategy effectively leverages the model’s attention capacity, allowing for more comprehensive utilization of contextual information.

We also observe a degradation phenomenon when the number of rotated prompts $N$ becomes too large (e.g., $N \geq 50$). In such cases, the model sometimes collapses to producing a single dominant prediction across inputs, losing discriminative power on certain classification tasks. This suggests that while EQR enhances robustness under low-resource settings, excessively large $N$ may dilute task-specific signals and harm decision diversity.

\section{Evaluation on Generation Tasks}
\label{sec:Generation Tasks}

To further assess the general applicability of DyVec, we evaluate it on six standard generation tasks using accuracy as the metric. As shown in Table~\ref{tab:linguistic task}, DyVec delivers competitive performance without increasing the prompt length, enabling more efficient inference. Although Few-shot ICL slightly outperforms DyVec on a few individual tasks, the overall results suggest that DyVec retains strong generalization ability beyond classification settings.

\section{Task Representation Transferability}
\label{sec:Task Representation Transferability}
To further evaluate the robustness of DyVec's latent modulation mechanism, we examine whether a fixed intervention position—learned on a particular dataset size—can generalize across activation vectors computed from different dataset sizes, Details can be found in Equation~\ref{eq:transfer}.

As shown in Table~\ref{tab:transfer}, the fixed intervention positions remain effective across a wide range of activation sources. This suggests that while the extracted activations may vary in granularity or representational strength, the optimal position for injecting modulation remains relatively stable within each model. This property enhances the flexibility of DyVec in real-world scenarios where activation extraction and position learning may occur under different data conditions.

\section{Inference methods}
\label{sec:Method}

Table~\ref{tab:Method} reports the best-performing intervention configurations $(\alpha, \beta)$ under the setting where the number of segmented positions $S$ equals the number of attention heads $H$. Each cell presents the configuration that achieved the lowest training loss for a specific model, task, and data size. These results reflect how the optimal intervention strategy can vary with both model architecture and the amount of training data, emphasizing the need for adaptive configuration in real-world applications.

\end{document}